\documentclass[11pt]{article}

\usepackage{acl}
\usepackage{graphicx}
\usepackage{amssymb} 
\usepackage{pifont}  
\usepackage{xcolor}  
\definecolor{ao}{rgb}{0.0, 0.5, 0.0} 
\definecolor{mediumgreen}{rgb}{0.56, 0.8, 0.56}
\usepackage{xcolor}
\definecolor{lightgreen}{rgb}{0.6, 1.0, 0.6}
\definecolor{darkgreen}{rgb}{0.0, 0.39, 0.0}
\definecolor{third}{HTML}{FFE5D9}
\definecolor{second}{HTML}{FFD7BA} 
\definecolor{best}{HTML}{FEC89A} 
\usepackage{booktabs}
\usepackage{times}
\usepackage{latexsym}
\usepackage{graphicx}
\usepackage{makecell}
\usepackage{booktabs}
\usepackage{amsmath}
\usepackage[T1]{fontenc}
\usepackage{comment}
\usepackage{caption} 
\usepackage{subcaption} 
\usepackage{tikz} 
\usepackage{pgfplots} 
\usepackage{fontawesome}
\usepackage{pifont} 
\usepackage{multirow}
\usepackage{hyperref}
\newcommand{\xmark}{\ding{55}}
\newcommand{\greencheck}{{\color{ao}\checkmark}} 
\newcommand{\redcross}{{\color{red}\xmark}} 
\usepackage[utf8]{inputenc}
\newcommand{\dataset}[1]{\text{GeoComp}}
\newcommand{\eval}[1]{\text{GeoEval}}
\usepackage{microtype}

\usepackage{inconsolata}
\usepackage[noblocks]{authblk}
\title{Geolocation with Real Human Gameplay Data:\\ A Large-Scale Dataset and Human-Like Reasoning Framework}
\author[1]{\textbf{Zirui Song}}
\author[2]{\textbf{Jingpu Yang}}
\author[2]{\textbf{Yuan Huang}}
\author[3]{\textbf{Jonathan Tonglet}}
\author[4]{\textbf{Zeyu Zhang}}

\author[5]{\protect\\ \textbf{Tao Cheng}} 
\author[6]{\textbf{Meng Fang}}
\author[1]{\textbf{Iryna Gurevych}}
\author[1]{\textbf{Xiuying Chen}}

\affil[ ]{\hspace{3em} \textsuperscript{1}MBZUAI \hspace{8em} \textsuperscript{2}Northeastern University}
\affil[ ]{\textsuperscript{3}TU Darmstadt and KU Leuven \hspace{3em} \textsuperscript{4}Australian National University} 

\affil[ ]{\textsuperscript{5}University College London \hspace{3em} \textsuperscript{6}University of Liverpool}

\begin{document}

\maketitle

\begin{abstract}
Geolocation aims to identify an image’s location and requires complex reasoning, playing an important role in navigation, monitoring, and cultural preservation. However, existing methods often yield coarse and non-interpretable predictions. A key challenge is the limited quality and scale of current geolocation datasets, which are typically small, automatically constructed, and suffer from noise and inconsistent difficulty.
To address these challenges, we introduce a comprehensive geolocation framework with three key components: \textbf{\textit{\dataset{}}}, a large-scale dataset; \textbf{\textit{GeoCoT}}, a novel reasoning method; and \textbf{\textit{GeoEval}}, designed to evaluate the correctness of the geolocation reasoning process.
At the core of this framework is \dataset{}, a large-scale dataset collected from a geolocation game platform involving 740K users over two years. 
It comprises 25 million entries of metadata and 2.7 million geo-tagged locations spanning much of the globe, with each location annotated thousands to tens of thousands of times by human users.
Building on this dataset, we propose Geographical Chain-of-Thought (GeoCoT), a multi-step reasoning framework designed to enhance the reasoning capabilities of Large Vision Models (LVMs) in geolocation tasks.
Finally, we demonstrate that GeoCoT significantly boosts performance by up to 25\% on classic geolocation metrics and by 9\% in reasoning quality as measured by GeoEval: \faGithub   ~\href{https://anonymous.4open.science/r/Geocomp-DDE1/README.md}{\textcolor{blue}{Github link}}.
\end{abstract}

\section{Introduction}

Geolocation, the task of determining an image’s geographical location, is crucial for applications like crime tracking, navigation, fact-checking, and cultural exploration~\cite{cheng2022,chalvatzaras2022}.
It involves interpreting contextual clues within an image, such as architectural styles, road signs, natural landscapes, and cultural markers.
Inferring location from such diverse indicators demands advanced reasoning, making geolocation a challenging task for both artificial models and human experts~\cite{khan2024debunking}.

\begin{table}[t!]
\centering
\resizebox{\columnwidth}{!}{ 
\begin{tabular}{@{}llcccc@{}}
\toprule
\multirow{2}{*}{\textbf{Dataset}} & \multirow{2}{*}{\textbf{Size}}   & \textbf{Geographic} &   \multirow{2}{*}{\textbf{Source}}& \textbf{Open} &\textbf{Human}   \\ 
&&\textbf{Coverage}&&\textbf{Access}& \textbf{Annotation} \\
\hline
Google-WS-15k
~\cite{clark2023we}
& 15k & \textcolor{lightgreen}{Global}&Map Service &\redcross &\redcross  \\
GMCP~\cite{zamir2014image}
& 105K & Local & Map Service & \redcross &\redcross  \\
StreetCLIP~\cite{haas2023learning} 
& 1M & Unknown& Map Service& \redcross&\redcross  \\
Im2GPS
~\cite{hays2008im2gps}
& 237 &Local & Web-Scraped& \greencheck &\redcross  \\ 
Im2GPS3K
~\cite{Im2GPS++YFCC4k+Im2GPS3k}
& 2997 &Local & Web-Scraped& \greencheck &\redcross  \\ 
YFCC4K
~\cite{Im2GPS++YFCC4k+Im2GPS3k}    
& 4536 & Local& Web-Scraped& \greencheck &\redcross  \\
YFCC26K
~\cite{YFCC-Val26k-Interpretable}
& 26k & Local & Web-Scraped& \greencheck &\redcross  \\
MP-16
~\cite{larson2017benchmarking}
& 4.7M & Local& Web-Scraped & \greencheck & \redcross \\
OSV-5M~\cite{astruc2024openstreetview}    
& 5.1M & \textcolor{mediumgreen}{Global} & Map Service & \greencheck & \redcross  \\ \hline
\textbf{\dataset{}} & 2.7M & \textcolor{darkgreen}{\textbf{Global}} & Map Service  & \greencheck  & \greencheck \\
\bottomrule
\end{tabular}%
}
\caption{Comparison of Existing Geolocation Datasets and GeoComp. 
``Local'' refers to city- or region-specific data, while ``Global'' spans multiple continents.  
Darker \textcolor{ao}{green} shades indicate broader geographic coverage. }
\label{table:dataset}
\vspace{-1em}
\end{table}

Significant effort has been devoted to solving the geolocation task~\cite{vo2017revisiting, PlaNet,zhu2022,muller2018}, but often at a coarse level of granularity and unlocalizability. 
The reason is potentially due to the lack of high-quality datasets. 
For example, Im2GPS3K contains up to 35\% non-localizable images~\cite{astruc2024openstreetview}, while the YFC100M dataset includes irrelevant data such as indoor photos and food images, which provide little to no locational information~\cite{YFCC-Val26k-Interpretable}. 
Additionally, many datasets are limited in size, with Georeasoner~\cite{ligeoreasoner} featuring only 3K images, thereby restricting the robustness and generalizability of geolocation models. 
A comparison of these datasets is shown in Table~\ref{table:dataset}.

To address the above obstacles, in this work, we leverage the contributions of hundreds of thousands of geolocation game enthusiasts who provide real user prediction annotations while playing the game. 
Specifically, we launched a free, public-benefit-oriented online geoguessing platform in June 2022.
A screenshot of the platform’s GUI is provided in Appendix~\ref {GUI}.
In each game, two players independently guess the location based on the same image and their own hints, with scores determined by the distance between their predictions and the ground-truth location.
The images are sourced from Google Maps, Baidu Maps, Tencent Maps, and Gaode Maps.
The platform offers multiple game modes, allowing users to either choose opponents or join random matchups.
As of December 2024, this platform has 740K users, 2.7M locations as unique geolocation tasks, and 25M human response records.
We name the collected dataset \dataset{}.
This rich and valuable dataset of real human responses enables us to evaluate task difficulty and filter out unreasonable cases.
For instance, some tasks are too easy, such as when the name of a shopping mall in a city is clearly visible in the image, enabling most users to answer correctly.
On the other hand, some tasks are highly challenging, where only a few users spend considerable time before providing accurate answers. 
Additionally, there are unreasonable tasks that contain no identifiable hints, making them unsolvable for all users despite significant effort.

Unlike previous approaches that address this task with a coarse level of granularity, we conduct a comprehensive evaluation of recent advanced LVMs on GeoComp, where the models are required to reason and predict the exact city of a given location.
Our findings reveal that this task poses a significant challenge for existing LVM models.
To address this, we introduce a Geographical Chain of Thought (GeoCoT) approach, which automatically guides the reasoning process through multi-step analysis of geographical cues, such as landmarks, environmental features, and spatial relationships.
For the evaluation of the reasoning process, we propose a set of articulated evaluation metrics, named as GeoEval including comparison with ground truth reasoning data and intrinsic evaluation.
The results demonstrate that our GeoCoT paradigm significantly improves geolocation accuracy.
It not only helps break down complex tasks into manageable reasoning steps but also enhances the interpretability of the inference process.

Our work makes key contributions to geolocation. 
First, we present \dataset{}, a large-scale, human-annotated geolocation dataset with over 2.7 million location images with corresponding location labels, and 25 million human player annotations, featuring diverse geographic regions, languages, and environmental contexts.
These annotations identify different difficulty geolocation cases and establish benchmarks to guide future advancements.
Second, we introduce the Geographical Chain of Thought (GeoCoT) framework, a multi-step reasoning approach that improves geolocation accuracy by leveraging geographical cues like landmarks, environmental features, and spatial relationships.  
Finally, through comprehensive evaluations involving human assessments and LLM inferences, we show that GeoCoT improves predictive performance by up to 25\% while enhancing interpretability.

\begin{figure*}[htb] 
    \centering
    \includegraphics[width=0.98\textwidth]{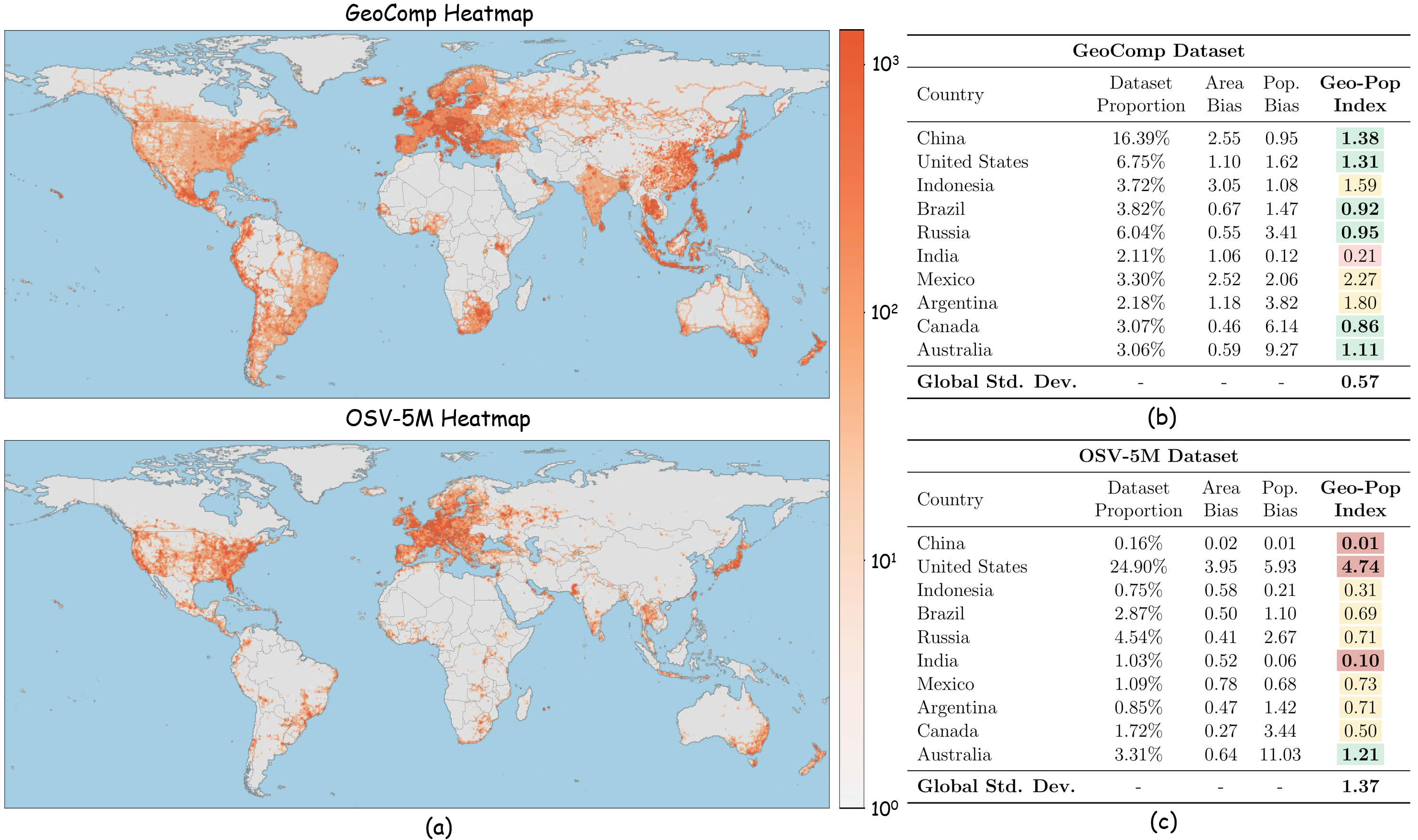} 
    \caption{
Comparison of data distribution between our GeoComp dataset and the existing SOTA OSV-5M dataset. (a) Global geographical heatmaps illustrating image density. (b–c) Country-level statistical breakdowns for GeoComp and OSV-5M, respectively.
The Geo-Pop Index, defined as the harmonic mean of Area Bias and Population Bias, is used to comprehensively assess distribution balance. Index values are color-coded to indicate balance levels:
\colorbox[HTML]{D4EFDF}{High Balance},
\colorbox[HTML]{FCF3CF}{Moderate Balance},
\colorbox[HTML]{FADBD8}{Imbalance}, and
\colorbox[HTML]{E6B0AA}{Critical Imbalance}.
}
    \label{fig:Geo}
    \vspace{-1em}
\end{figure*}

\section{Related work}

\subsection{Image Geolocation Task}
\label{related1}
Image geolocation refers to determining the corresponding location of a given image, a crucial task in computer vision~\cite{zhu2023difftraj,zhu2023synmob,zhu2024controltraj}, spatial data mining~\cite{zhao2017incorporating,zhao2016exploring,zhang2023promptst,han2023mitigating}, and GeoAI~\cite{zhao2017modeling,zhao2022multi,zhang2023mlpst,zhang2023autostl}.
Previous research in image geolocalization could be primarily classified into two approaches: classification-based methods and retrieval-based methods.
(1) \textit{Classification-based methods} partition most regions of the Earth into multiple grid cells.
Models are trained to classify each image into the correct cell~\cite{clark2023, pramanick2022, muller2018, seo2018, weyand2016}. 
The center coordinates of each cell are used as the predicted values. 
However, due to the limited number of cells, the granularity of the predicted values is coarse, preventing precise predictions.
(2) \textit{Retrieval-based methods} establish a database of geographic images with GPS coordinates. 
For a given input image, these methods retrieve the most similar image from the dataset and use its coordinates as the predicted location~\cite{zhu2022,muller2018, zhang2023,workman2015wide,liu2019lending,Zhou_2024}. 
However, constructing a comprehensive global-level image database is clearly impractical.

\subsection{Geolocation Dataset}

Existing geolocation datasets primarily originate from web-scraped or street-view images that have not been human-validated, raising concerns about their quality for effectively evaluating geolocation capabilities. 
For instance, datasets derived from web scraping, such as YFCC100M~\cite{theiner2022} and Im2GPS3K~\cite{vo2017revisiting}, include a significant proportion of images depicting food, art, pets, and personal photographs. 
These types of images are often weakly localizable or entirely non-localizable~\cite{YFCC-Val26k-Interpretable}.
Street-view datasets also face limitations, such as restricted geographic coverage~\cite{astruc2024openstreetview}; for example, \cite{mirowski2019streetlearn-navigation}'s work includes data from only three cities in the United States.
Furthermore, dataset collection processes often introduce biases. 
For instance, some commonly used platforms are inaccessible in certain countries, resulting in uneven geographic representation. 
Additionally, the difficulty of individual geolocation tasks varies widely within these datasets, but this aspect has not been comprehensively evaluated. 
For example, images taken at prominent landmarks are relatively easy to geolocate, while others offer no clear hints and are highly challenging~\cite{astruc2024openstreetview}.
These limitations undermine the reliability of current geolocation benchmarks.

\subsection{Large Vision Language Models}

Numerous studies have been conducted on various aspects of LVMs, encompassing structural design \cite{liu2024visual, cai2023benchlmm, liu2024sphinx}, data construction \cite{laion2023gpt4v, zhao2024cobra}, training strategies \cite{zhao2023mamo,mckinzie2024mm1, lu2024deepseek}, evaluations~\cite{bithel2023evaluating}, and the development of lightweight LVMs \cite{ zhu2024comprehensive}.
Additionally, the robust capabilities of LVMs have been applied to other fields, such as medical image understanding \cite{tinyllava, pmc-vqa,zhang2024universal} and document parsing \cite{ye2023mplug, liu2024textmonkey}. 
However, the reasoning capabilities of LVMs in geolocation tasks remain underexplored. 
One of the primary reasons for this limitation is the lack of high-reasoning-value geographic data.

\section{Data Overview}
\subsection{Geolocation Competition}

Inspired by geoguessr website, we developed a free geolocation game platform that tracks participants' competition histories.
Unlike most geolocation websites, including Geoguessr, which rely solely on samples from Google Street View, our platform integrates Baidu Maps, Tencent Maps, and Gaode Maps to address coverage gaps in regions like mainland China, ensuring broader global accessibility.
Users can choose specific opponents or engage in random matches. 
Each match consists of multiple questions, and each user is initially assigned a “vitality score.” 
Users mark their predicted location on a map, and the system evaluates accuracy based on the surface distance between the predicted point and the ground truth. 
Larger errors result in greater deductions from the user's vitality score. 
The user with the higher vitality score at the end of the match is declared the winner.
To ensure predictions are human-generated rather than machine-generated, users must register with a phone number, enabling tracking of individual activities. 
Using this platform, we collected \dataset{}, a comprehensive dataset covering 1,000 days of user competition.

\subsection{Dataset Statistics and Geographic Balance}
To comprehensively evaluate the geographical distribution quality of our dataset, we conduct a comparative analysis with the existing SOTA dataset, OSV-5M~\cite{astruc2024openstreetview}.
Figure \ref{fig:Geo}(a) visualizes the global density of the dataset via geographical heatmaps. Compared to OSV-5M, which exhibits a skewed distribution towards Western regions, \dataset{} demonstrates a significantly more comprehensive coverage.
First, \dataset{} successfully mitigates the data sparsity in under-represented areas, achieving dense coverage in East Asia and South America. Second, our improved global balance does not sacrifice the density of data in well-covered areas. As evidenced by the higher heatmap intensity, our dataset maintains superior sampling density even in Western regions (e.g., Europe and North America) compared to OSV-5M. This dual advantage, expanding global reach while intensifying local density, ensures that GeoComp is both spatially balanced and densely distributed.

To further validate the distribution quality beyond visual inspection, we propose the \textit{Area Bias} and \textit{Population Bias} ratios, calculated as the quotient of a country's proportion in the dataset to its respective global share of land area or population.
To synthesize spatial and population factors, we define the \textit{Geo-Pop} Index as the harmonic mean of Area Bias and Population Bias, where 1.0 indicates perfect parity.
We categorize distribution quality into four levels based on the magnitude of deviation:High Balance ($0.8\text{--}1.4$) denotes optimal alignment with real-world shares; Moderate Balance ($0.3\text{--}0.8$ or $1.4\text{--}3.0$) and Imbalance ($0.1\text{--}0.3$ or $3.0\text{--}4.5$) reflect increasing degrees of skew; while Critical Imbalance ($<0.1$ or $>4.5$) identifies extreme outliers that are either under-represented by over $10\times$ or over-sampled by more than $4.5\times$.
The statistical breakdown reveals a severe polarization in OSV-5M. As shown in Figure \ref{fig:Geo}(c), it exhibits Critical Imbalance in major nations: the United States is disproportionately over-represented (accounting for $24.90\%$ of the data with an Index of $4.74$), while China is severely under-represented (only $0.16\%$, Index $0.01$).
In contrast, \dataset{} effectively mitigates these imbalances, achieving a status of High Balance across the mainstream country (Figure \ref{fig:Geo}(b)). For instance, the representation of China is restored to a reasonable $16.39\%$ (Index $1.38$), and the United States is normalized to $6.75\%$ (Index $1.31$), reflecting a distribution more aligned with real-world geography and population.
Overall, the majority of countries in \dataset{} fall within the High Balance range, demonstrating a stable distribution (Global Std. Dev. of $0.57$). Conversely, OSV-5M is predominantly characterized by Moderate Balance and Critical Imbalance, resulting in a significantly higher deviation of $1.37$.

\begin{figure*}[htb]
    \centering
    \includegraphics[width=1\linewidth]{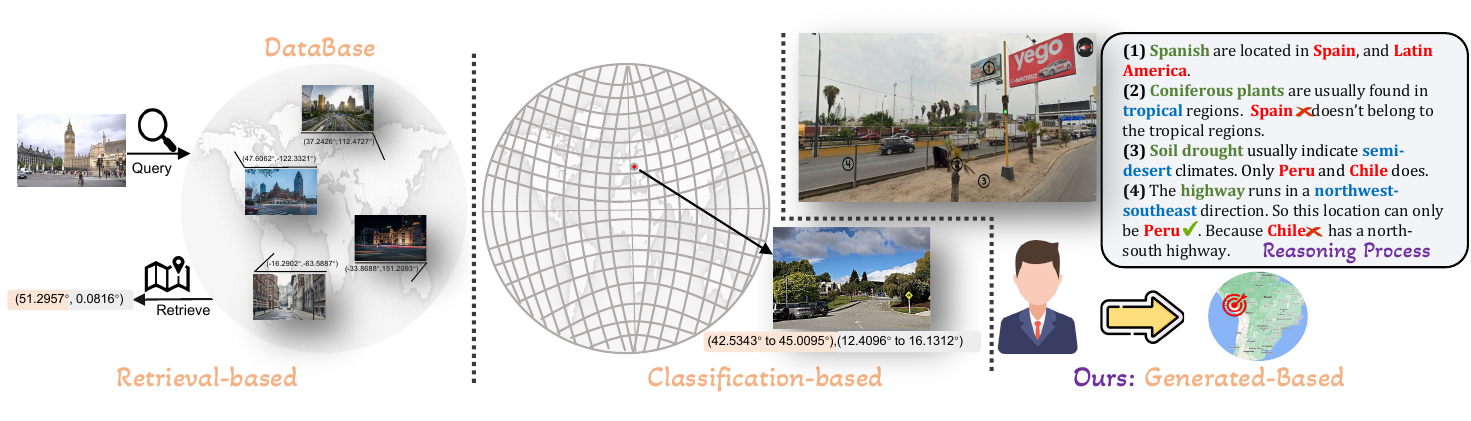}
    \caption{Comparison of previous geolocation tasks and our proposed paradigm: while previous works focused on coarse-grained predictions limited by dataset quality, our generation and reasoning-based method enables fine-grained city-level predictions.}
    \label{fig:comp}
    \vspace{-1em}
\end{figure*}

\subsection{Human Annotations}

\begin{figure}
    \centering
    \includegraphics[width=1\linewidth]{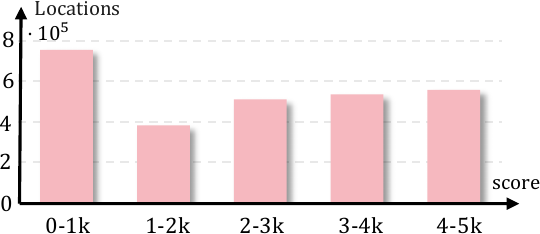}
    \caption{\textcolor{black}{Number of locations distributed across different average score intervals.}}
    \label{fig:score_dist}
    \vspace{-1em}
\end{figure}

Distinct from previous works, \dataset{} includes rich human gameplay data, serving as a vital benchmark for evaluating geolocation difficulty and image localizability. 
We analyze performance using the GeoGuessr scoring mechanism, where a user's score $S$ decays exponentially with the distance error $d$. 
The score is calculated as $S = \lfloor \exp(-d/s_d) \times 5000 \rfloor$, where $s_d$ is a scale factor derived from the map's maximum distance. A perfect prediction ($d=0$) yields 5,000 points.
To establish a reliable difficulty benchmark,  we visualized the dataset for the annotated locations and calculated the average score for each location.
The resulting distribution is shown in Figure~\ref{fig:score_dist}.

As illustrated in the histogram, the dataset exhibits a broad spectrum of difficulty levels.
The 0--1,000 score range constitutes the largest portion of the data, accounting for approximately 27.6\%.
This high concentration of low scores reveals a substantial subset of hard samples, which are typically located in remote or rural areas and lack explicit semantic cues such as text signs.
For these locations, accurate localization requires fine-grained reasoning over implicit environmental features, including vegetation patterns, soil texture, and architectural styles.
The poor human performance in this range suggests that average players generally lack the specialized domain knowledge needed to interpret these subtle cues, which often results in near-random guessing.
This characteristic highlights the opportunity for models to surpass human performance by effectively learning high-entropy environmental representations.
Overall, this distribution confirms that \dataset{} serves as a robust benchmark, providing sufficient challenging samples to stress state-of-the-art models while retaining a substantial volume of human-solvable data to support meaningful geographic feature learning.



\begin{table*}[htb]
\centering
\resizebox{0.9\textwidth}{!}{%
\begin{tabular}{lccccccccc}
\toprule
Model & \multicolumn{3}{c}{City} & \multicolumn{3}{c}{Country} & \multicolumn{3}{c}{Continent}\\
& Accuracy$\uparrow$ & Recall$\uparrow$ & F1$\uparrow$ & Accuracy$\uparrow$ & Recall$\uparrow$ & F1$\uparrow$ & Accuracy$\uparrow$ & Recall$\uparrow$ & F1$\uparrow$ \\
\midrule
LLaVA-1.6   & 0.002  & 0.001  & 0.002 & 0.041 & 0.019  & 0.028  & 0.175 & 0.067 & 0.056\\
LLama-3.2-Vision & 0.081 & 0.037 & 0.030 & \colorbox{best}{0.630} & \colorbox{second}{0.199} & \colorbox{second}{0.217} & \colorbox{second}{0.866} & \colorbox{second}{0.643} & \colorbox{third}{0.639}  \\
Qwen-VL   & 0.016  & 0.013  & 0.014  & 0.069  & 0.042  & 0.070  & 0.130 & 0.115  & 0.077\\
GeoCLIP & 0.018 & 0.007 & 0.008 & 0.550 & \colorbox{third}{0.197} & 0.204 & \colorbox{best}{0.872} & \colorbox{best}{0.746} & \colorbox{best}{0.731} \\
GeoReasoner & 0.018  & 0.014  & 0.012  & 0.092  & 0.053  & 0.085  & 0.208 & 0.161 & 0.144\\
Kimi-latest & 0.027 & 0.023 & 0.024 & 0.135 & 0.112 & 0.156 & 0.315 & 0.148 & 0.220 \\
Kimi-latest(CoT) & 0.028 & 0.029 & 0.029 & 0.139 & 0.092 & 0.124 & 0.331 & 0.154 & 0.210 \\
Kimi-latest(GeoCoT) & 0.038 & 0.035 & 0.029 & 0.239 & 0.192 & 0.224 & 0.361 & 0.168 & 0.232 \\
GPT-4o & \colorbox{third}{0.092} & \colorbox{third}{0.048} & \colorbox{third}{0.044} & 0.615 & 0.188 & 0.208 & 0.807 & 0.468 & 0.487 \\
GPT-4o(CoT) & \colorbox{second}{0.094} & \colorbox{second}{0.052} & \colorbox{second}{0.042} & \colorbox{second}{0.623} & 0.194 & \colorbox{second}{0.212} & 0.819 & 0.430 & 0.449 \\
 GeoCoT  & \colorbox{best}{\textbf{0.118}} & \colorbox{best}{\textbf{0.089}} & \colorbox{best}{\textbf{0.086}} & \colorbox{best}{\textbf{0.640}} & \colorbox{best}{\textbf{0.260}} & \colorbox{best}{\textbf{0.291}} & \colorbox{third}{0.862} & \colorbox{third}{0.638} & \colorbox{second}{0.646}  \\
\bottomrule 
\end{tabular}%
} 
\caption{Comparison of Precision, Recall and F1 scores in country-level and city-level geolocation. The scores are represented as follows: \colorbox{best}{best}, \colorbox{second}{second}, and \colorbox{third}{third}. 
Numbers in \textbf{bold} mean that the improvement to the best baseline is statistically significant (a two-tailed paired t-test with p-value \textless 0.05).}
\label{tab:main}
\vspace{-1em}
\end{table*}

\section{Geographic Chain of Thought}

\subsection{Rethinking Geolocation Task}

As discussed in \S~\ref{related1}, the geolocation task has traditionally relied on classification-based~\cite{clark2023, pramanick2022, weyand2016} and retrieval-based methods~\cite{zhu2022, zhang2023}, as shown in Figure~\ref{fig:comp}.
While these approaches have advanced the field, they face significant limitations in precision and scalability, prompting a rethinking of the task.
Inspired by how humans gradually narrow down locations from broad to fine-grained observations~\cite{luo2022g3}, we propose a new geolocation paradigm: predicting geographic locations through a step-by-step reasoning process.
Unlike traditional approaches limited by grid-based classification and exhaustive databases, our model generates natural language reasoning, guiding it to the final predicted city. 
To implement this paradigm, we introduce GeoCoT (Geographic Chain-of-Thought), a framework designed for both interpretability and accuracy.

\subsection{GeoCoT Deisgn}
Our design of GeoCoT is inspired by how humans intuitively approach geolocation—progressing from broad to fine-grained analysis.
Rather than relying on generic step-by-step reasoning like standard CoT prompting, GeoCoT mimics the human process: starting with macro-level cues (e.g., climate, terrain), then narrowing down to country, city, and finally micro-level details to guide the model through interpretable geographic reasoning.

Concretely, GeoCoT operates in five sequential stages:
\textit{1. Continental or Climate Zone Identification.}
The process begins with identifying broad regions using natural features like mountains, vegetation, or soil, narrowing the scope to a continent or climate zone.
\textit{2. Country-Level Localization.}
Cultural markers, language on signs, and architectural styles are analyzed to refine predictions to the country level.
\textit{3. City-Level Refinement Using Infrastructure.}
Street elements, such as driving direction, bollards, and license plate colors, are used to locate specific cities or regions.
\textit{4. Landmark-Based Verification.}
Features like fire hydrants, guideposts, and street signs help validate and further refine the predicted location.
\textit{5. Fine-Grained Micro-Level Validation.}
Finally, subtle details such as sidewalk patterns and clothing styles confirm precise localization at a city or neighborhood level.
These five reasoning steps are formulated as a single, structured prompt and jointly fed into the LVM, which directly generates the final predicted location.
Detailed prompts can be found in Appendix~\ref{details}.

It is important to note that GeoCoT does not require any concrete knowledge about the specific features of locations. 
Instead, it offers reasoning tutorials designed to help LVMs identify geographic clues by leveraging their existing knowledge.
Our inference time and efficiency analysis in Appendix~\ref{appendix:inference_time} shows that GeoCoT incurs only about a 10\% increase in inference latency compared to standard CoT.
We further provide an ablation study in Appendix~\ref{appendix:ablation}.

\section{Experiments}

\subsection{Setting}
We selected 500 geo-tagged locations with high inferential value from the dataset to serve as a test set, using a stratified sampling method across continents to ensure balanced geographic distribution.
This number is larger than in previous works \cite{liu2023visual, guan2024hallusionbench}, which typically include only a few dozen case studies to examine the reasoning process.
Specifically, we selected 20 mainstream countries across six continents as representative samples and extracted tasks with an average player score of around 3,000 out of 5,000 for annotation.
This test set has been publicly released on GitHub.

\subsection{Baselines}

We compare GeoCoT against strong baselines representing recent geolocation advancements. For general open-source VLMs, we evaluate LLaVA-1.6~\cite{liu2023improvedllava}, Llama-3.2-vision~\cite{meta2024llama}, and Qwen2.5-VL~\cite{qwen-lm}, also include geolocation-specific models GeoCLIP~\cite{vivanco2024geoclip} and GeoReasoner~\cite{ligeoreasoner}. Furthermore, we assess closed-source performance using GPT-4o~\cite{gpt-4o}, Kimi k2 ~\cite{team2025kimi} and also utilizing chain-of-thought reasoning~\cite{wei2022chain}. All models are evaluated using the same input format and test set to ensure a fair comparison.

\subsection{Overall Performance Evaluation of GeoCoT}
\label{sec:5.2}

We evaluate geolocation performance from two aspects: first, location prediction compared with the ground truth at various levels; and second, the direct calculation of the Earth's surface distance.  
We present the location prediction performance in Table~\ref{tab:main}, evaluated across three levels: city, country, and continent.
Performance is measured using \textit{accuracy}, which calculates the proportion of correct predictions out of all predictions; \textit{recall}, which determines the proportion of true positive predictions out of all actual positive cases; and the \textit{F1} score, which balances precision and recall to provide their harmonic mean.  

The results reveal several key observations. 
First, open-source LVMs such as LLaMA-3.2-Vision achieve competitive performance, performing on par with GPT-4o and GPT-4o (CoT), demonstrating their effectiveness in location prediction tasks. 
Second, performance varies across different levels of granularity. 
While GPT-4o (CoT) ranks second at the city level, it underperforms at the country level, highlighting the importance of multi-level evaluation to fully assess a model’s geolocation reasoning ability. 
Finally, our model, GeoCoT, consistently achieves top performance across all nine metrics and three levels, demonstrating its robustness and adaptability in geolocation tasks.
Additionally, GeoCLIP surpasses GPT-4o at the continent level, which can be attributed to its pretraining on image-GPS pairs, making it particularly well-suited for coarse-grained geolocation tasks. 
Coarse-grained continent-level predictions typically require less detailed local knowledge and instead rely on broader geographic cues, such as climate, landscapes, and cultural markers. 
However, GeoCLIP performs poorly at finer granularities like city levels, suggesting that it lacks a strong capability for geographic reasoning beyond direct visual features.

\begin{table}[h]
\centering
\small
\begin{tabular}{lccc}
\toprule
\multirow{2}{*}{\textbf{Model}} & \text{Street} & \text{City} & \text{Country} \\ 
&\text{1km}&\text{25km}&\text{750km} \\ \hline
LLaVA-1.6 &0.006 &0.020 &0.082 \\
Llama-3.2-Vision &0.018 &0.104 &0.638 \\
Qwen-VL  &0.004 &0.014 &0.090 \\
GeoCLIP&0.035&0.077&0.625  \\
GeoReasoner&0.010&0.020&0.128  \\
GPT-4o & 0.045 & 0.147 & 0.678 \\
GPT-4o(CoT) & 0.047 & 0.151 & 0.701 \\
GeoCoT&  \textbf{0.073}  & \textbf{0.157} & \textbf{0.711} \\ 
\bottomrule
\end{tabular}
\caption{Accuracy of different models on geolocation tasks at various scales.
Numbers in \textbf{bold} mean that the improvement to the best baseline is statistically significant (a two-tailed paired t-test with p-value \textless 0.05).}
\label{tab:model_accuracy}
\vspace{-1em}
\end{table}

Next, in Table~\ref{tab:model_accuracy}, we present the accuracy of each model by measuring the geographic distance between the predicted city and the ground truth. The metrics represent the proportion of predictions within three distance thresholds: Street (1 km), City (25 km), and Country (750 km). 
Higher values indicate better performance, with stricter thresholds assessing fine-grained localization and larger thresholds evaluating coarse-level accuracy.
The results show that GPT-4o and Llama-3.2-vision outperform the dedicated large-scale model GeoCLIP for geolocation, even under finer-grained evaluation settings. 
For example, at the street-level threshold, GPT-4o achieves 0.045 compared to GeoCLIP’s 0.035, and at the city-level threshold, GPT-4o scores 0.147, nearly double GeoCLIP’s 0.077.
Moreover, our proposed GeoCoT paradigm demonstrates even greater improvements.
At the street level, GeoCoT achieves 0.073, significantly outperforming both GeoCLIP (0.035) and GPT-4o (0.045). 
Similarly, at the city level, GeoCoT achieves 0.157, and at the country level (750 km), it achieves 0.711, the highest among all models.
These results highlight GeoCoT’s strong performance and the potential of its reasoning framework for geolocation tasks.
We further provide an ablation study on GeoCoT in Appendix~\ref{appendix:ablation}.

\begin{table}[htb]
\centering
\resizebox{0.48\textwidth}{!}{
\begin{tabular}{lcccccc}
\toprule
\multirow{3}{*}{\textbf{Model}} & &\textbf{Im2GPS} & &  &\textbf{Im2GPS3K} & \\
& \text{Street} & \text{City} & \text{Country} & Street & City & Country \\ 
&\text{1km}&\text{25km}&\text{750km} & 1km & 25km  & 750km  \\ \hline
LLaVA-1.6 &0.04 &0.18 & 0.39&0.03 &0.14 & 0.32 \\ 
Llama-3.2-Vision &0.09 &0.37 & 0.65&0.07 &0.27 & 0.52 \\
Qwen-VL &0.04 &0.21 & 0.37&0.04 &0.15 & 0.26 \\
GeoCLIP     & 0.17  & 0.41 & 0.77 & 0.13 &0.32&0.67  \\
GeoReasoner &0.05 &0.19 & 0.33&0.04 &0.15 & 0.26 \\
PlaNet &0.08 &0.25 & 0.54&0.09 &0.25 & 0.48 \\ 
GPT-4o & 0.13 & 0.47 & 0.74 & 0.14 & 0.40 & 0.66 \\
GPT-4o(CoT) & 0.16 & 0.49 & 0.77 & 0.14 & 0.45 & 0.69 \\
GeoCoT& \textbf{0.22} & \textbf{0.55} & \textbf{0.83} & \textbf{0.15} & \textbf{0.46} & \textbf{0.74} \\ 
\bottomrule
\end{tabular}
}
\caption{Performance comparison of GeoCoT and state-of-the-art geolocation models on traditional benchmarks.
Numbers in \textbf{bold} mean that the improvement to the best baseline is statistically significant (a two-tailed paired t-test with p-value \textless 0.05).}
\label{tab:traditional}
\vspace{-1em}
\end{table}

\subsection{Generalizability Evaluation}
\label{section:general_eval}

Even though our dataset is more comprehensive and human-annotated, we are also interested in evaluating how our model performs on traditional geolocation datasets to provide a more thorough comparison.
Hence, we select two existing benchmark datasets, Im2GPS~\cite{hays2008im2gps} and Im2GPS3K~\cite{Im2GPS++YFCC4k+Im2GPS3k}, due to their popularity and widespread use in geolocation tasks as standard benchmarks for evaluating model performance.
We present the performance results in Table~\ref{tab:traditional}.
We observe that state-of-the-art geolocation models, such as GeoCLIP, perform well on traditional geolocation tasks, surpassing GPT-4o and coming close to our model, GeoCoT.
However, this is in contrast to the results shown in Table~\ref{tab:main}, where GeoCLIP significantly underperforms GPT-4o on fine-grained city- and country-level geolocation tasks.
This discrepancy suggests that these baseline models may be overfitting to the specific datasets they were trained on, lacking the generalization ability required for more diverse or fine-grained geolocation challenges.

\subsection{GeoEval: Reference-Based Evaluation}
\label{sec:5.4}
Beyond evaluating overall task performance, we focus on analyzing the reasoning process of GeoCoT, which emulates a human-like reasoning approach. 
To establish a reference for this evaluation, three gaming enthusiasts collaboratively constructed reasoning processes for the same 500 cases based on geo-tagged locations. We designated these as the reasoning ground truth (a human-annotated example can be found in Appendix~\ref{human}).
These GT annotations serve as a benchmark within our evaluation framework, GeoEval. 
The evaluation process utilizes GPT-based assessment through GPTScore~\cite{fu2023gptscore} and prompt-based scoring.


Our prompt-based scoring includes four dimensions (scored 0-5; detailed prompts are on Github).
First, \textit{\textbf{completeness of feature extraction (CE)}} evaluates whether all key clues in the GT are comprehensively covered, ensuring reasoning relies on sufficient factual evidence.
Second, \textit{\textbf{accuracy of feature extraction (AE)}} measures the correctness of identified attributes to prevent reasoning deviations caused by misidentified features.
Third, \textit{\textbf{accuracy of reasoning and cue correspondence (AC)}} assesses whether conclusions derived from extracted cues are reasonable and consistent with the GT logic.
Finally, \textit{\textbf{logical coherence of reasoning (LC)}} evaluates the consistency, flow, and adherence to common sense within the reasoning chain, ensuring the reliability of the conclusion.

\begin{table}[ht]
\centering
\small
\resizebox{0.48\textwidth}{!}{%
\begin{tabular}{lcccccc}
\toprule
Model & Similarity & \multicolumn{4}{c}{GeoEval} \\
& {\text{GPTScore}} & {\text{CE}} & \text{AE} & \text{AC} & \text{LC} \\
\midrule
LLaVA-1.6 & 0.478 & 1.262 & 1.271 & 1.446 & 1.490 \\
Llama-3.2-Vision & 0.566 & 2.203 & 2.386 & 2.558 & 2.721 \\
Qwen-VL & 0.371 & 1.231 & 1.255 & 1.453 & 1.484 \\
GeoReasoner & 0.424 & 1.421 & 1.533 & 1.719 & 2.038 \\
GPT-4o & 0.613 & 2.320 & 2.891 & 2.809 & 3.143 \\
GPT-4o(CoT) & 0.663 & 2.462 & 3.136 & 3.156 & 3.540 \\
GeoCoT & \textbf{0.728} & \textbf{2.690} & \textbf{3.538} & \textbf{3.696} & \textbf{3.945} \\
\bottomrule
\end{tabular}%
}
\caption{Evaluation of GeoCoT's reasoning process using ground truth-based metrics within the GeoEval framework. 
Numbers in \textbf{bold} mean that the improvement to the best baseline is statistically significant (a two-tailed paired t-test with p-value \textless 0.05).}
\label{tab:reasoning_eval}
\vspace{-1em}
\end{table}

The experimental results in Table~\ref{tab:reasoning_eval} highlight the significant advantages of GeoCoT compared to baseline models across all evaluation metrics.
GeoCoT achieves the highest GPTScore of 0.728, outperforming GPT-4o (CoT) (0.663) and LLaVA-1.6 (0.478), demonstrating its superior alignment with human-constructed reasoning processes. 
In terms of feature extraction, GeoCoT achieves a CE score of 2.690 and an AE score of 3.538, significantly surpassing GPT-4o (CoT) and the dedicated GeoReasoner model. 
These results clearly demonstrate that GeoCoT not only captures key information more comprehensively but also maintains a more accurate and logically coherent reasoning process compared to both reasoning-based models and traditional baselines like GeoReasoner.

\subsection{Intrinsic Evaluation of GeoCoT Reasoning}

\label{sec:5.3}

We begin with a ground truth-based evaluation, comparing GeoCoT’s reasoning to human-authored processes to assess its alignment with established reasoning patterns.
To complement this, we conduct an intrinsic evaluation focused on hallucination errors, coherence, and robustness without relying on external references.
Given the need for multimodal judgment, this evaluation is performed by human annotators.
\begin{table}[h]
\vspace{-1em}
\small
\centering
\begin{tabular}{lccc}
\toprule
\multirow{2}{*}{\textbf{Model}} & {OH} & {FH} & {AH} \\
               & Count$\downarrow$ 
               & Count$\downarrow$  
               & Count$\downarrow$  \\ \midrule
GeoReasoner    &237                     &151                                            
               &203                                                     \\
GPT-4o         &43                     &4                              
               &35                                                                 \\

GeoCoT         &5                     &1                                
               &18                                \\ 
\bottomrule
\end{tabular}
\caption{Hallucination Evaluation on Reasoning Data.}
\label{tab:reasoning_value}
\vspace{-1em}
\end{table}

Following previous work on assessing hallucinations in terms of objects, attributes, and relationships~\cite{li2023evaluating,sun2023aligning}, we evaluate the quality of synthetic data across three key dimensions: (1) \textbf{Object Hallucination (OH)}: assesses whether the synthetic data includes objects that do not exist in the image. Object Hallucination evaluates the extent to which synthetic data introduces fictional elements. 
(2) \textbf{Fact Hallucination (FH)} measures the accuracy of factual information within the synthetic data. Fact Hallucination occurs when the synthetic data contains facts, figures, or other information that is incorrect or not supported by the original data.   (3) \textbf{Attribution Hallucination (AH)} evaluates whether the synthetic data incorrectly attributes properties, characteristics, or relations to entities or objects. 
To quantify hallucinations, each detected error is counted as one instance in the corresponding dimension.
To evaluate these dimensions, we invited 2 human annotators with professional backgrounds in geographic reasoning and data validation to assess GPT-4o, GeoReasoner, and our proposed GeoCoT model. 
These three baselines provide textual reasoning processes across 1,500 evaluated cases. 
The results, shown in Table~\ref{tab:reasoning_value}, indicate the number of errors in each dimension, demonstrating that GeoCoT significantly reduces hallucination errors compared to the other models.
The inter-annotator agreement, measured by Cohen's Kappa, is 0.82 for OH, 0.79 for FH, and 0.85 for AH, indicating substantial agreement across all dimensions. The detailed reasoning case studies are listed in Appendix~\ref{appendix:case_study}.

\section{Conclusion}

In this work, we advance the field of image geolocation by introducing \dataset{}, the largest human-annotated dataset to date, comprising 2.7 million locations and 25 million real-world gameplay records.
Leveraging this rich data, we propose GeoCoT, a geographical chain-of-thought framework that mimics human reasoning to enhance LVM capabilities.
Comprehensive evaluations using our novel GeoEval metric demonstrate that GeoCoT significantly outperforms state-of-the-art baselines in both accuracy and interpretability.

\section*{Limitations}

Despite the advancements presented, our work has three main limitations.
First, although \dataset{} is globally diverse, geographic coverage remains uneven, with sparse representation in regions like Africa and parts of Oceania due to limited map service availability.
Nevertheless, compared to existing large-scale geolocation benchmarks (e.g., OSV-5M), \dataset{} substantially mitigates extreme regional bias and achieves a markedly more balanced global distribution, as demonstrated by the Geo-Pop Index analysis.
Second, GeoCoT relies on computationally intensive LVM (e.g., GPT-4o). The multi-step reasoning process increases inference latency and cost compared to lightweight, embedding-based retrieval methods, potentially limiting real-time deployment.
While this overhead is inherent to CoT-style reasoning, our empirical analysis (Appendix~\ref{appendix:inference_time}) indicates that GeoCoT does not incur substantially higher latency than standard CoT prompting across different models.
Finally, while GeoCoT excels at city-level localization through reasoning, it may struggle with precise meter-level coordinate prediction in non-urban environments where semantic cues (text, landmarks) are absent, an area where retrieval-based methods still hold an advantage.
This limitation stems from the intrinsic reliance of reasoning-based methods on interpretable visual cues; in such low-semantic settings, direct visual matching remains more effective, and we view the integration of reasoning-based and retrieval-based paradigms as a promising direction for future work.

\section*{Data Ethics}
The creation and release of our dataset adhere to stringent ethical standards to ensure the privacy and well-being of all contributors. 
We have conducted rigorous anonymization of the dataset to protect user privacy. 
All personally identifiable information, such as usernames, email addresses, and IP addresses, has been permanently removed. 
Only non-identifiable behavioral data, such as prediction outcomes and timestamps, are retained. 
The dataset originates from user participation on our open-source geolocation game platform. 
Users were informed during the registration process that their activity data might be used for research purposes. 
This ensures transparency in data collection and maintains user trust. 
We have explicitly designed the dataset for research purposes, with the sole intention of advancing geolocation and related artificial intelligence technologies.
Importantly, our dataset does not include the images directly but instead provides links to images hosted on platforms such as Google Maps or Baidu Maps, which can be accessed through their official APIs.
We are committed to ensuring the responsible use of this dataset. Researchers accessing the data must agree to a data usage agreement that prohibits unethical or illegal use. 
\bibliography{custom}
\clearpage
\newpage

\appendix

\section{Data Collection Platform User Interface}
\label{GUI}
\begin{figure}[]
    \centering
    \includegraphics[width=0.8\linewidth]{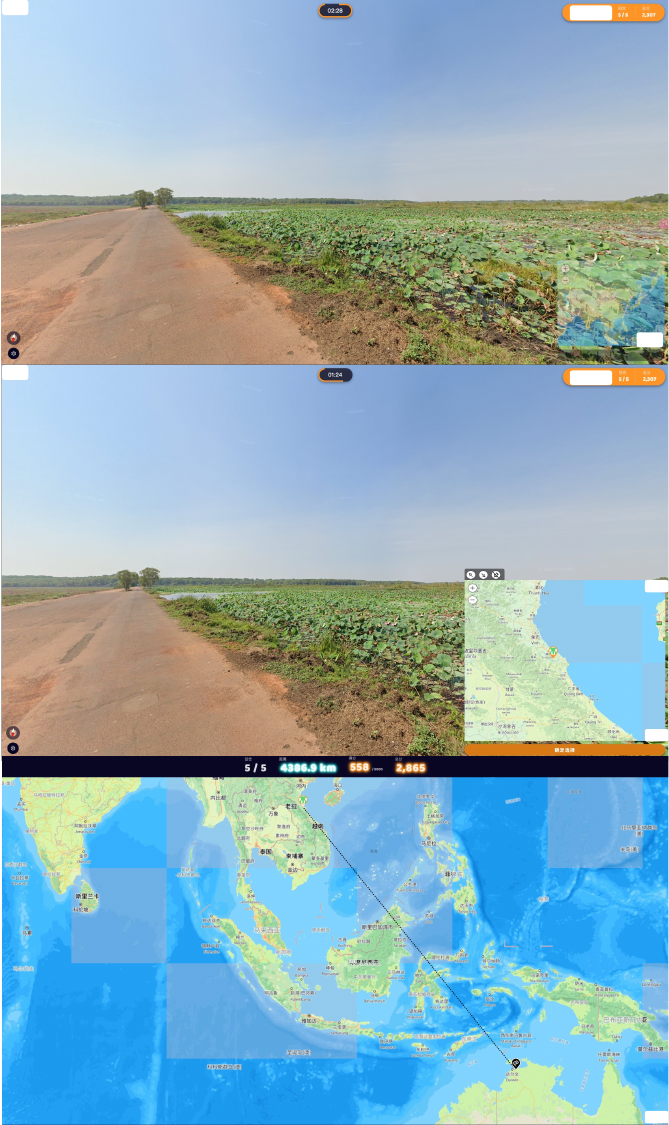}
    \caption{UI of Gameplay. UI components that could potentially compromise the double-blind review policy were masked.}
    \label{fig:UI}
\end{figure}

To comply with the double-blind review policy, we did not include the URL of our active website in the paper. Instead, we presented selected interface screenshots of the website in Figure ~\ref{fig:UI} while obscuring any elements that could potentially compromise the anonymity required by the policy. 

\section{Detail of GeoCoT}
\label{details}
We present the detailed prompt of our GeoCoT process below:

\textit{$\bullet$ \textbf{Question1:} Are there prominent natural features, such as specific types of \textcolor{cyan}{vegetation}, \textcolor{cyan}{landforms} (e.g., \textcolor{orange}{mountains}, \textcolor{orange}{hills}, \textcolor{orange}{plains}), or \textcolor{cyan}{soil characteristics}, that provide clues about the geographical\textcolor{green}{region}?
 \noindent $\bullet$ \textbf{Question2:} Are there any culturally, historically, or architecturally significant \textcolor{cyan}{landmarks}, \textcolor{cyan}{buildings}, or \textcolor{cyan}{structures}, or are there any \textcolor{cyan}{inscriptions} or \textcolor{cyan}{signs} in a specific
    language or script that could help determine the \textcolor{green}{country} \textcolor{green}{or region}?
\noindent $\bullet$  \textbf{Question3:} Are there distinctive road-related features, such as \textcolor{cyan}{traffic direction} (e.g., \textcolor{orange}{left-hand or right-hand driving}), specific types of \textcolor{cyan}{bollards}, unique utility \textcolor{cyan}{pole designs}, or \textcolor{cyan}{license plate}colors and styles, which \textcolor{green}{countries} are known to have these characteristics?
 \noindent $\bullet$ \textbf{Question4:} Are there observable \textcolor{cyan}{urban} or \textcolor{cyan}{rural markers} (e.g., \textcolor{orange}{street signs}, \textcolor{orange}{fire hydrants guideposts}) , or other
    \textcolor{cyan}{infrastructure} elements, that can provide more specific information about the \textcolor{green}{country or city}?
\noindent $\bullet$  \textbf{Question5:} Are there identifiable patterns in \textcolor{cyan}{sidewalks} (e.g., \textcolor{orange}{tile shapes}, \textcolor{orange}{colors}, or \textcolor{orange}{arrangements}), \textcolor{cyan}{clothing styles} worn by people, or other culturally specific details that can help narrow down the \textcolor{green}{city or area}?}

\begin{figure*}[h]
    \centering
    \includegraphics[width=0.95\linewidth]{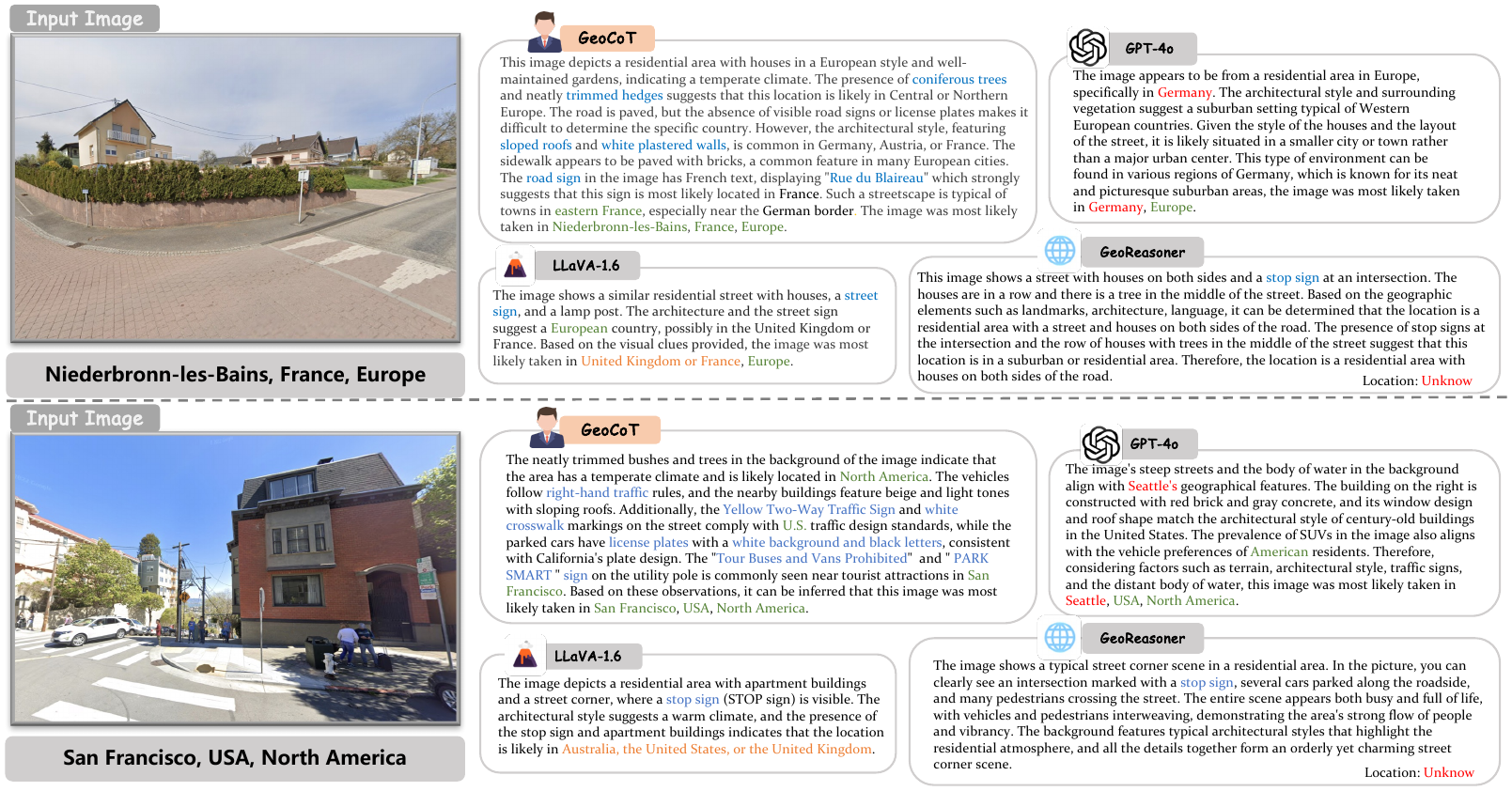}
    \caption{Qualitative comparison of LLaVA, GPT4o, and GeoReasoner. Clues are shown in \textcolor[HTML]{2f6eba}{blue}, correct predictions in \textcolor[HTML]{628443}{green}, incorrect in \textcolor[HTML]{ea3323}{red}, and vague/uncertain guesses in \textcolor[HTML]{de8344}{orange}.}
    \label{fig:Case_study}
\vspace{-1em}
\end{figure*}

\textit{Let's think step by step. Based on the question I provided, locate the location of the picture as accurately as possible. Identify the continent, country, and city, and summarize it into a paragraph. 
For example: the presence of tropical rainforests, palm trees, and red soil indicates a tropical climate... Signs in Thai, right-side traffic, and traditional Thai architecture further suggest it is in Thailand... Combining these clues, this image was likely taken in a city in \textcolor{red!70!black}{Bangkok, Thailand, Asia.}}

Here, \textcolor{cyan}{cyan} highlights potential clues within the image to help the model infer geographic locations. \textcolor{green}{Green} defines the geographic scope inferred from the clues, such as a region, country, or city. \textcolor{orange}{Orange} provides detailed descriptions of the cyan clues, enhancing the model's understanding. \textcolor{red!70!black}{Red} specifies the expected output format, including city, country, and continent.

\section{Human Annotation Example}
\label{human}
Below we show an example of human annotated ground truth to demonstrate the annotation process, criteria, and the reasoning behind the annotations, where clues are shown in \textcolor[HTML]{2f6eba}{blue}, correct predictions in \textcolor[HTML]{628443}{green}.

\textit{The image shows a rural residential area with dense trees and expansive green lawns. The terrain is flat, and the \textcolor[HTML]{2f6eba}{soil is reddish-brown}, which matches the temperate climate of central Europe, particularly rural areas of France. The architectural style of the house is distinctive: a \textcolor[HTML]{2f6eba}{red-tiled sloped roof, yellow walls, and solar panels}, reflecting the region's focus on renewable energy, a common feature in French countryside homes. The \textcolor[HTML]{2f6eba}{red mailbox} at the gate is a hallmark of rural French residences. The design of the fences and modern \textcolor[HTML]{2f6eba}{gates aligns} with typical styles in the French countryside. The house design and surrounding natural environment suggest a rural European region. Based on the architectural style, natural landscape, and street elements, the image was most likely taken in \textcolor[HTML]{628443}{Aumont, France, Europe}.}

\section{Case Study}
\label{appendix:case_study}

Figure~\ref{fig:Case_study} presents qualitative comparisons. In the first case, GeoCoT correctly identifies the location in France by integrating architectural and environmental clues, whereas baselines struggle with specific regional markers. In the second case, GeoCoT leverages text and traffic signs for precise city-level localization, while baselines like LLaVA and GPT-4o fail to capture fine-grained details.

\section{Inference Time and Efficiency Analysis}
\label{appendix:inference_time}

To address concerns regarding computational overhead, we conducted a comprehensive performance analysis on our test set. It is important to note that a key advantage of our GeoCoT framework is its efficiency in deployment: it does not require any additional model training or fine-tuning. Instead, our contribution focuses on engineering a structured chain-of-thought prompt that elicits advanced reasoning capabilities from off-the-shelf Large Vision Models. Furthermore, we benchmark our framework against GLOBE \cite{Globe}, a recent reasoning-driven approach that reinforces image geo-localization through policy optimization and enhanced visual-cue reasoning.
Table~\ref{tab:inference_time} presents the average number of tokens generated and the corresponding inference time for the base model (GPT-4o), a standard Chain-of-Thought (CoT) approach, and our proposed GeoCoT framework.

\begin{table}[h]
\centering
\resizebox{0.48\textwidth}{!}{%
\begin{tabular}{lcc}
\toprule
\textbf{Model} & \textbf{Avg. Generated Tokens} & \textbf{Avg. Inference Time (s)} \\
\midrule
GLOBE & 397.74 & 8.13 \\
GPT-4o & 89.76 & 5.62 \\
GPT-4o (Standard CoT) & 141.58 & 7.57 \\
\textbf{GeoCoT (Ours)} & \textbf{173.28} & \textbf{8.88} \\
\bottomrule
\end{tabular}%
}
\caption{Comparison of inference time and token generation across different methods. GeoCoT introduces only a marginal increase in latency compared to standard CoT.}
\label{tab:inference_time}
\end{table}

The results indicate that GeoCoT introduces a modest and acceptable latency overhead,
with approximately 1.3 seconds over standard CoT and 3.2 seconds over the base model.
 This marginal increase in response time is a direct trade-off for the more comprehensive, multi-step reasoning process required for precise geolocation. Given the significant gains in localization accuracy demonstrated in the main experiments, we consider this computational cost to be acceptable for practical applications.

\section{Ablation Study on Reasoning Steps}
\label{appendix:ablation}

\begin{table*}[t]
\centering
\resizebox{\textwidth}{!}{%
\begin{tabular}{lccccccccc}
\toprule
\multirow{2}{*}{\textbf{Configuration}} & \multicolumn{3}{c}{\textbf{City}} & \multicolumn{3}{c}{\textbf{Country}} & \multicolumn{3}{c}{\textbf{Continent}} \\
\cmidrule(lr){2-4} \cmidrule(lr){5-7} \cmidrule(lr){8-10}
& Accuracy & Recall & F1 & Accuracy & Recall & F1 & Accuracy & Recall & F1 \\
\midrule
Step1 & 0.113 & 0.098 & 0.076 & 0.551 & 0.133 & 0.151 & 0.776 & 0.443 & 0.468 \\
Step2 & 0.103 & 0.086 & 0.074 & 0.510 & 0.157 & 0.181 & 0.755 & 0.438 & 0.464 \\
Step3 & 0.102 & 0.073 & 0.076 & 0.490 & 0.142 & 0.165 & 0.673 & 0.334 & 0.380 \\
Step4 & 0.082 & 0.078 & 0.069 & 0.469 & 0.122 & 0.142 & 0.735 & 0.346 & 0.369 \\
Step5 & 0.102 & 0.088 & 0.078 & 0.551 & 0.151 & 0.173 & 0.776 & 0.338 & 0.363 \\
\midrule
Step1+2 & 0.112 & 0.083 & 0.086 & 0.571 & 0.196 & 0.190 & 0.795 & 0.443 & 0.468 \\
Step1+2+3 & 0.114 & 0.102 & 0.089 & 0.592 & 0.216 & 0.192 & 0.816 & 0.453 & 0.474 \\
Step1+2+3+4 & 0.112 & 0.089 & 0.078 & 0.604 & 0.231 & 0.293 & 0.824 & 0.550 & 0.572 \\
\textbf{Step1+2+3+4+5 (GeoCoT)} & \textbf{0.118} & \textbf{0.089} & \textbf{0.086} & \textbf{0.640} & \textbf{0.260} & \textbf{0.291} & \textbf{0.862} & \textbf{0.638} & \textbf{0.646} \\
\bottomrule
\end{tabular}%
}
\caption{Ablation study on the cumulative effect of different reasoning steps within the GeoCoT framework. The results demonstrate that the combination of all five steps yields the best performance.}
\label{tab:ablation_steps}
\end{table*}

To assess the contribution of each reasoning stage within our framework, we conducted an ablation study on the reasoning steps. The detailed results are presented in Table~\ref{tab:ablation_steps}.

The experimental results indicate that single-step reasoning (e.g., relying solely on climate or infrastructure) fails to fully leverage the advantages of GeoCoT. This limitation is highlighted by the marginal improvements observed across all three evaluation metrics (City, Country, and Continent). As shown in the table, the performance steadily improves as more reasoning steps are integrated. Consequently, the full efficacy of our GeoCoT framework is only realized through the synergistic integration of multi-step reasoning (Step 1 through Step 5), which achieves the highest accuracy and F1 scores across all granularities.

\end{document}